\documentclass[letterpaper, 10pt, conference]{ieeeconf}
\IEEEoverridecommandlockouts    

\usepackage{cite}
\usepackage{booktabs}
\usepackage{multirow}
\usepackage{lipsum}
\usepackage{amsmath}
\usepackage{amssymb}
\usepackage[ruled,vlined]{algorithm2e}
\usepackage{graphicx}
\graphicspath{{./Figures/}}
\usepackage[pagebackref,breaklinks,colorlinks]{hyperref}

\title{Multi-modal Video Representation Alignment for Robust Self-supervised Driver Distraction Detection}

\author{David J. Lerch$^{1,2}$ \thanks{$^{1}$ Fraunhofer IOSB, Karlsruhe, Germany {\tt\small \{firstname.lastname\}@iosb.fraunhofer.de}}, Livien Majer$^{1,2}$, Zeyun Zhong$^{1,2}$, Manuel Martin$^{1}$, \\ Frederik Diederichs$^{1}$, Rainer Stiefelhagen$^{2}$ \thanks{$^{2}$ Karlsruhe Institute of Technology (KIT), Karlsruhe, Germany {\tt\small \{firstname.lastname\}@kit.edu}}}

\begin{document}
	
	\maketitle

    \begin{abstract}
Robust self-supervised learning of multi-modal video representations is critical for real-world applications such as driver distraction detection, where multiple sensors provide complementary but noisy signals. Conventional contrastive objectives, such as InfoNCE, assume all negatives are equally informative and all positives are reliable. However, this assumption is frequently violated in multi-modal data due to viewpoint changes, occlusions, or semantic overlap across modalities. In this work, we propose a novel framework for \emph{multi-modal global alignment} that addresses these challenges by jointly modeling faulty negatives and unreliable or faulty positives. We introduce soft targets derived from cycle-consistency scores to relax the hard-negative assumption, and a weighting mechanism based on similarity distributions to mitigate the impact of noisy or faulty positives. Our approach extends traditional pairwise alignment to a principled global multi-modal setting, aggregating alignment information across all modality pairs. We evaluate our method on the Drive\&Act dataset, demonstrating that it consistently outperforms both pairwise and existing global alignment baselines across RGB, IR, Depth, and Skeleton modalities. Cross-view ablation studies further show strong generalization to unseen camera perspectives, highlighting the robustness of our representations. Overall, our framework provides a scalable and effective solution for self-supervised global multi-modal representation learning, enabling reliable driver distraction detection and pioneering in real-world multi-modal video understanding. Our code will be published on GitHub.
\end{abstract}
\section{Introduction}
\label{sec:intro}
Driver distraction has emerged as a leading contributor to road traffic accidents worldwide, posing a persistent threat to global road safety. According to the European Commission’s 2025 Road safety thematic report, the exact number of road crashes caused by distracted drivers remains unknown~\cite{EC_ERSO_DriverDistraction_2025}. Austrian crash data shows that in recent years the proportion of distraction and inattention as the main crash cause increased to 32.7\% of injury crashes and 28\% of fatal crashes in 2024~\cite{statistik_at_strassenverkehrsunfaelle_2023} while data from the United States from 2021 show distraction involved in 8\% of fatal crashes, 14\% of injury crashes, and 13\% of all police-reported motor vehicle traffic crashes~\cite{NHTSADistractedDriving2023}. Naturalistic driving studies further suggest higher prevalence, with Dingus et al.~\cite{Dingus2016Distracted} finding that 68.3\% of 905 crashes involved observable distraction. These figures underscore the urgent need for advanced detection technologies amid rising in-vehicle distractions from smartphones, infotainment systems, and secondary tasks

\begin{figure}[t]
  \centering
  \includegraphics[width=0.9\linewidth]{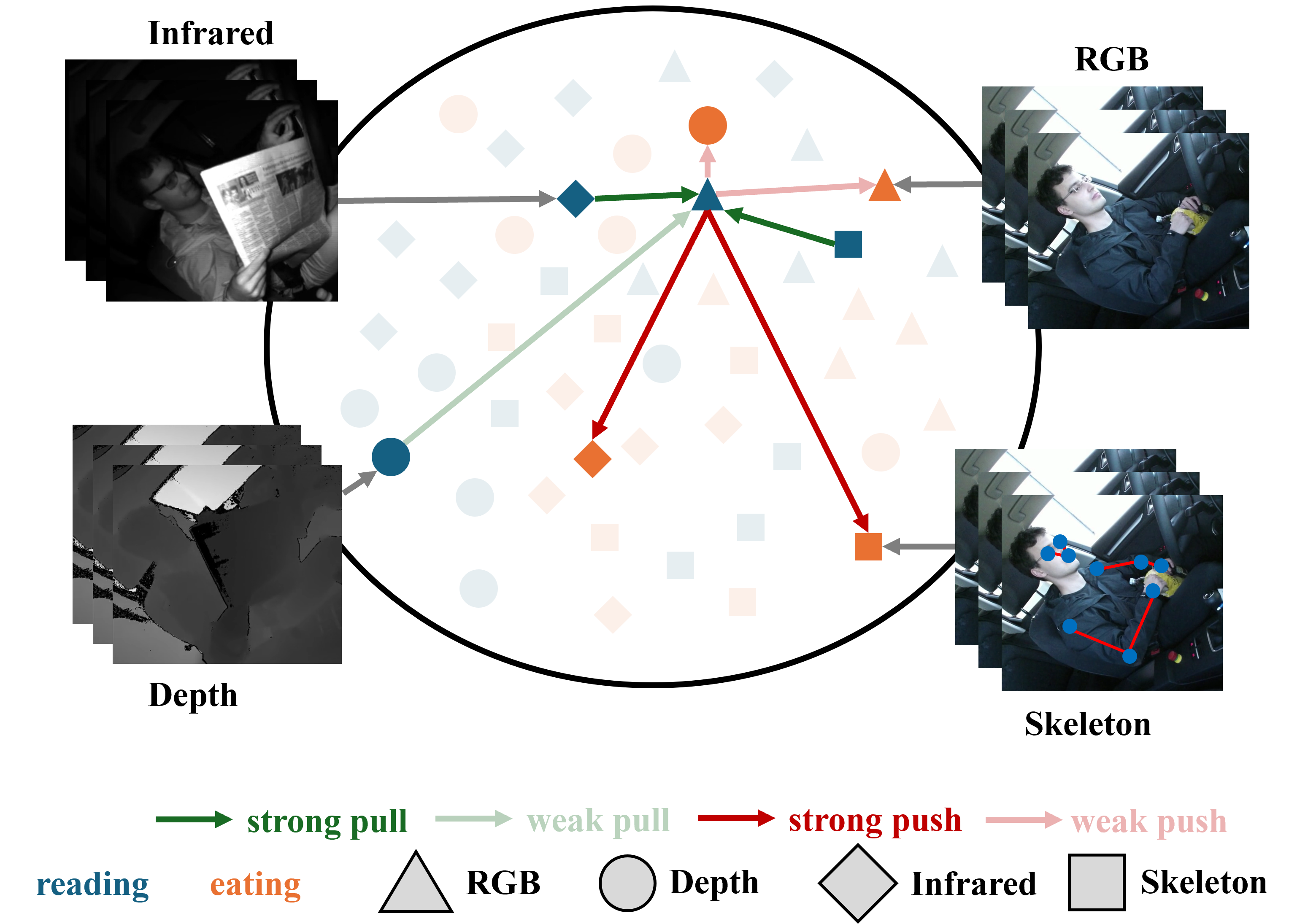}
   \caption{Overview of our robust multi-modal alignment for a RGB video sample of the activity eating. Positive pairs are pulled together, while negative pairs are pushed apart. In our approach we preserve the semantics by letting semantically dissimilar positive pairs staying apart~(weak pull) and semantically similar negative pairs be close together~(weak push).}
   \label{fig:concept}
\end{figure}

Driver Monitoring Systems (DMS), integral to Advanced Driver Assistance Systems (ADAS), aim to detect distraction, drowsiness, and fatigue in real time to enable timely interventions. Traditional single-modality approaches like physiological signals (e.g., heart rate, EEG), vehicle telemetry (e.g., steering, lane deviation), or vision-based cues (e.g., gaze, head pose) each face limitations: physiological methods are intrusive and costly, vehicle data is susceptible to noise and context-dependence, and vision struggles with lighting or occlusion~\cite{li_multimodal_framework_fatigue_driving_2026}. Multi-modal systems, fusing different sensor modalities, offer greater robustness by capturing complementary behavioral indicators, enhancing accuracy in diverse real-world scenarios. In our approach we use different visual modalities and different views to learn representations robust to lighting and occlusions.

Despite these advantages, multi-modal driver distraction detection grapples with significant challenges, particularly data imbalance. Naturalistic datasets exhibit severe class skew, wherein non-distracted driving predominates, while safety-critical distractions (e.g., texting, phone use) are rare and long-tailed. This imbalance biases deep learning models toward majority classes, yielding poor recall for minority distraction events with out-sized crash risk. Fusion has been shown to exacerbate issues through modality misalignment, missing data, and heterogeneous noise, demanding specialized strategies like temporal synchronization, attention-based integration, and imbalance mitigation (e.g., cost-sensitive losses~\cite{lin2017focal, morgado2021robust}, resampling~\cite{kang2020decoupling}, synthetic oversampling~\cite{ijerph18147534, Lerch2024self}).

Recent advances in self-supervised learning (SSL)~\cite{dino_caron2021emerging} and contrastive methods~\cite{CLIP} address these gaps by pretraining on vast unlabeled multi-modal data, yielding generalizable features that reduce annotation needs. SSL encoders excel at capturing semantic driver behaviors without exhaustive labeling, pairing effectively with imbalance-aware techniques for deployable DMS.

This paper tackles driver distraction detection in multi-modal, imbalanced datasets, proposing a novel framework for adaptive fusion and long-tailed learning. Key contributions include:
\begin{enumerate}
    \item We propose global modality alignment with a semantic-aware loss for driver distraction detection
    \item We show that our global modality alignment~(GMA) outperforms state-of-the-art local modality alignment~(LMA) methods on the driver activity recognition benchmark dataset Drive\&Act~\cite{martin2019drive_and_act_2019_iccv}
    \item Cross-view evaluation demonstrates that encoders aligned with our proposed GMA method show strong generalization to novel views.
\end{enumerate}

    \section{Related Work}

\subsection{Multi-modal Alignment for Driver Distraction Detection}

The detection of driver distraction has traditionally relied on supervised learning approaches, which leverage labeled datasets such as the State Farm~\cite{statefarm2016} dataset, Driver Monitoring Dataset~\cite{ortega2020dmd} and the Drive\&Act~\cite{martin2019drive_and_act_2019_iccv} dataset. Architectures like CNN-based VGG variants~\cite{VGG_GAP_zhou2016learning}, Drive-Net~\cite{59_MIM_Sup_majdi2018drive}, InceptionV3~\cite{Inception_V3_szegedy2016rethinking} and ResNet~\cite{Resnet_50_he2016deep} ensembles achieved notable accuracy in controlled experimental conditions~\cite{61_MIM_Sup_dhakate2020distracted}. However, the reliance on large, annotated datasets remains a limiting factor, as labeled driving data is both costly and constrained in reflecting real-world variability . Consequently, there has been a growing shift towards self-supervised representation learning, enabling models to learn from raw, unlabeled sensory data and generalize better across environmental conditions.

Recent advances in self-supervised driver distraction detection have demonstrated the potential of pretext tasks such as masked image modeling and contrastive learning to generate meaningful representations. Zhang et al.~\cite{masked_image_modeling_dd_zhang2023novel} proposed SL-DDBD, which integrates the Swin Transformer~\cite{swin_transformer_paper_liu2021swin} within a masked modeling framework to capture high-level semantic relationships from visual cues. Similarly, Li et al.~\cite{MIM_21_li2021driver} introduced an unsupervised deep learning model utilizing contrastive loss over feature projections, reducing label dependency while improving cross-domain adaptability. 

These methodologies highlight the scalability and robustness of self-supervised representation learning for visual understanding in vehicular contexts.

\subsection{Multi-modal Learning in Action and Driver Understanding}

Concurrent with the emergence of driver monitoring, multi-modal learning has come to the fore as a prevailing paradigm in human action recognition. This emerging field integrates features from a variety of modalities, including RGB, infrared, depth, and 3D skeletons, to leverage the complementary information offered by these diverse sources. Multi-stream fusion schemes effectively capture both appearance-based and structural cues, as observed in RGB-depth fusion networks and generalized models such as M-Mixer~\cite{lee2023mmixer}, which combine RGB, IR, and depth modalities. These integrations proved crucial for improving recognition accuracy under varied environmental conditions.

In the context of driver state and distraction analysis, multi-modality can extend beyond visual data to include gaze tracking, skeleton keypoints, steering wheel angles, and vehicular telemetry~\cite{kazemi2025evaluating, rezaei2025driver}. Prior supervised fusion models~\cite{rezaei2025driver} employed co-learning strategies for RGB–skeleton alignment, but such techniques often depend on synchronized labeled samples. 

In contrast, recent frameworks~\cite{ImageBind, languagebind, cmvra, zhu2024part} inspired by vision-language models reveal opportunities for aligning rich sensory streams in a self-supervised, contrastive learning manner. These models establish common embedding spaces by encouraging consistency among different modalities without explicit annotation, enabling cross-modal retrieval and generalization.

For our work we employ a multi-modal alignment framework that is trained self-supervised without the requirement for manual labels.

\subsection{Multi-modal Contrastive Alignment}

Contrastive learning has become a foundational principle for cross-modal representation alignment, with InfoNCE~\cite{infoNCE} serving as a primary objective for modality matching. ImageBind~\cite{ImageBind} introduced a bidirectional InfoNCE formulation to connect images with diverse sensory inputs (e.g., audio, depth, motion), demonstrating scalable cross-modal reasoning. Therefore modalities are aligned pairwise to the RGB image anchor. However, pairwise alignment is inherently limited, as it assumes a strict one-to-one correspondence between modalities and fails to exploit higher-order relationships.

To address this, multi-modal extensions have been proposed like the Cross-Modal Visual Representation Alignment (CMVRA) framework~\cite{cmvra}, which generalizes bidirectional InfoNCE to jointly align M modalities via averaged pairwise losses. This enables the model to capture shared semantics across all modalities simultaneously while preserving modality specific features. Such alignment is crucial for driver distraction detection, where diverse sensory signals, such as visual cues, body posture, gaze patterns, and vehicle dynamics, carry correlated but non-redundant information.

While most multi-modal methods use InfoNCE loss for alignment, we leverage a more robust loss specifically designed for noisy data and long tailed data distribution.
\begin{figure*}[t]
  \centering
  \includegraphics[width=0.9\linewidth]{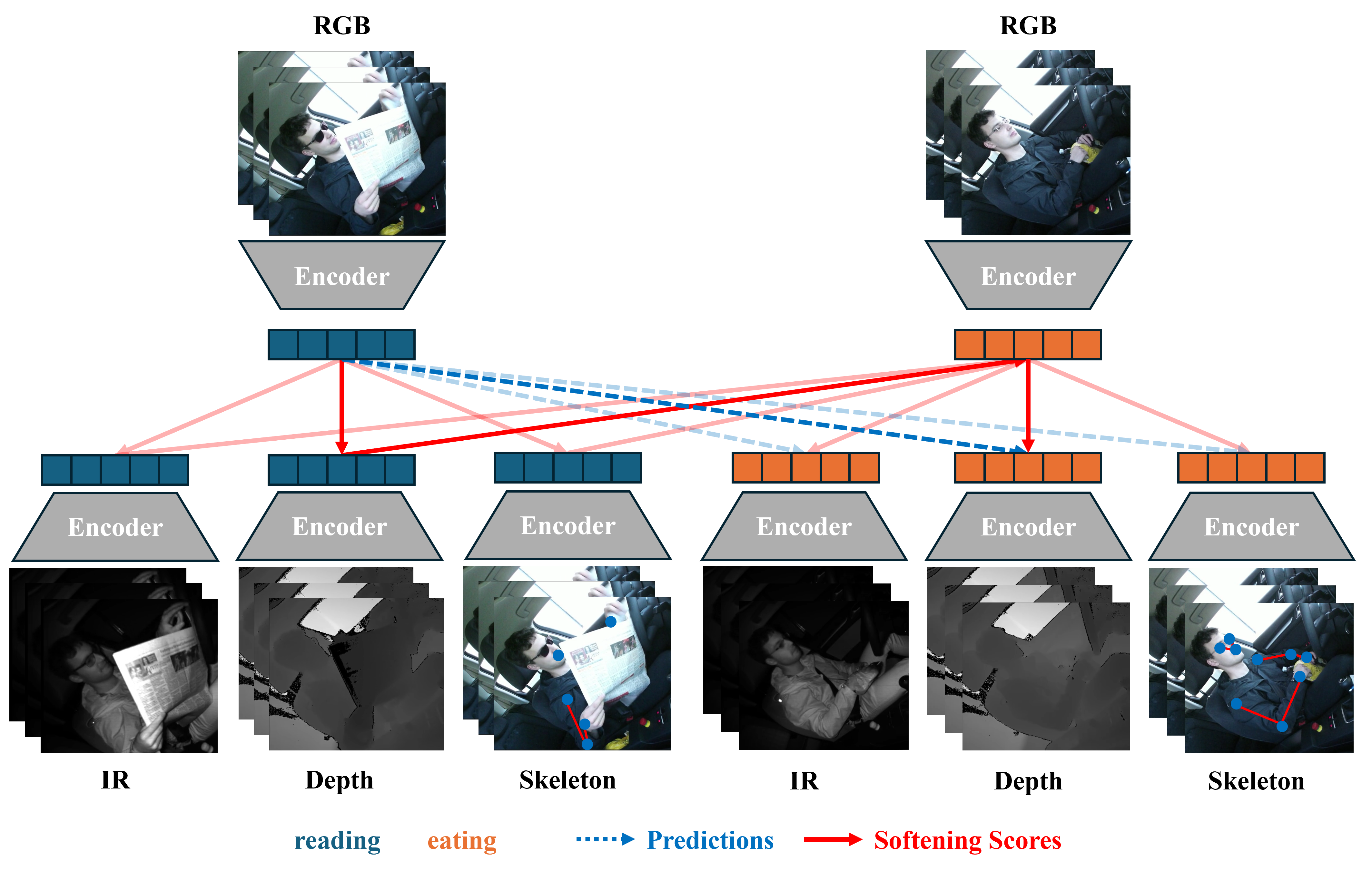}
   \caption{Principle of our cycle consistency across multiple modalities. For each modality we calculate the cycle consistent soft targets to all other modalities. We use the cycle consistent soft targets to calculate the loss across all available modalities.}
   \label{fig:principle}
\end{figure*}

\subsection{Robust Contrastive Objectives for Noisy Multi-modal Data}

Despite their success, contrastive methods remain sensitive to false positives and negatives, especially in real-world settings where different modalities may contain asynchronous or noisy correspondences. Additionally to noisy and asynchronous correspondences, Bhardwaj et al.~\cite{Lerch2025self} introduced clustering to handle data imbalanced data and reduce biases is self-supervised learning. Methods such as False Negative Cancellation~\cite{huynh2022boosting}, Incremental False Negative Detection~\cite{chen2022incremental}, and Prototypical Contrastive Learning~\cite{li2021prototypical} mitigate these limitations by refining the loss landscape through dynamic sample weighting or prototype-based clustering. In multi-modal contexts, soft target mechanisms~\cite{morgado2021robust, chen2024vision_language_skeleton} further relax the rigid one-to-one assumption by introducing graded similarity measures and cycle-consistent training signals. These techniques improve semantic consistency across modalities while maintaining stability during joint training.

Building on these developments, our approach introduces a global multi-modal soft-weighted loss that extends soft-target concepts for more than two input modalities. Our method is specifically tailored for driver distraction detection. By aggregating feature statistics and leveraging global consistency across all available modalities, the proposed loss formulation ensures robust multi-modal alignment under conditions of noisy and imbalanced data.
    \section{Methods}

\subsection{Dataprocessing}
For the Drive\&Act dataset, we construct a unified multi-modal data pipeline that enables joint learning from RGB, IR, depth, and 3D skeleton information for driver activity recognition. Each dataset element corresponds to a short, temporally localized clip that is consistently defined across all modalities. Visual clips are temporally synchronized by enforcing a fixed frame rate and a fixed number of frames per clip, ensuring that all modalities describe the same temporal window.

The multi-modal representation is organized at the clip level: for each annotated temporal segment, we extract a fixed-length sequence of frames per modality. Visual modalities (RGB, IR, depth) are treated as image sequences with three channels, while skeleton sequences are treated as structured image-like inputs where each 3D joint acts as a spatial “patch.” This design yields a homogeneous interface to the model across heterogeneous input types. At training and evaluation time, each sample is represented as a dictionary containing the tensors of all initialized modalities together with the corresponding class label, enabling flexible ablation over subsets of modalities.
These preprocessing steps ensure consistent temporal and spatial alignment across modalities 
and provide the input embeddings for our method.

\subsection{Alignment Settings}

We focus on aligning all modalities jointly and call this global modality alignment (GMA) in contrast to pairwise or local modality alignment (LMA). In the following we focus on GMA across RGB, depth, IR, and skeleton modalities, 
as this setting provides the most challenging scenario and allows a thorough evaluation of our method. The RGB encoder remains frozen to provide a stable anchor for all other modalities, following the CMVRA~\cite{cmvra} design.

Building on the CMVRA framework~\cite{cmvra}, we propose a robust multi-modal alignment loss that addresses two key limitations of CMVRA: the handling of faulty positives, and faulty negatives~(see Figure~\ref{fig:concept}). Our proposed method combines a multi-modal weighting mechanism inspired by Robust xID~\cite{morgado2021robust} and a soft-target strategy based on self-similarity~\cite{chen2024vision_language_skeleton}.

\subsection{Preliminaries and Notation}

Let $M$ denote the number of modalities and $B$ the batch size.
For each modality $m \in \{1, \dots, M\}$, we denote the batch of
$d$-dimensional embeddings as

\begin{equation}
Z^{(m)} = 
\begin{bmatrix}
z^{(m)}_1 \\
\vdots \\
z^{(m)}_B
\end{bmatrix}
\in \mathbb{R}^{B \times d},
\end{equation}

where $z^{(m)}_b \in \mathbb{R}^d$ is the embedding of sample $b$
from modality $m$.
All embeddings are $\ell_2$-normalized before similarity computation.


\subsection{Symmetric Cross-modal Alignment}

For a modality pair $(m,n)$, we compute the similarity matrix

\begin{equation}
S^{(m,n)} = Z^{(m)} (Z^{(n)})^\top
\in \mathbb{R}^{B \times B}.
\end{equation}

Each entry $S^{(m,n)}_{b,k}$ corresponds to the dot product
between sample $b$ from modality $m$ and sample $k$ from modality $n$.

Temperature-scaled logits are defined as

\begin{equation}
L^{(m \to n)} = \frac{S^{(m,n)}}{\tau},
\qquad
L^{(n \to m)} = \left(L^{(m \to n)}\right)^\top.
\end{equation}

Row-wise softmax yields conditional probability matrices

\begin{equation}
P^{(m \to n)}_{b,k}
=
\frac{
\exp\left(L^{(m \to n)}_{b,k}\right)
}{
\sum_{\ell=1}^{B}
\exp\left(L^{(m \to n)}_{b,\ell}\right)
},
\end{equation}

where $\tau > 0$ denotes the temperature parameter

This formulation is fully symmetrical, and anchor samples are not a prerequisite.
Each row defines a conditional distribution over batch elements.

\subsection{Soft Targets for Faulty Negatives}

To mitigate the effect of false negatives, we introduce soft target matrices. The principle of our multi-modal soft-targets is demonstrated in Figure~\ref{fig:principle}. 
Let $I \in \mathbb{R}^{B \times B}$ denote the identity matrix.
Given a soft similarity-based target matrix
$T^{(m \to n)}_{\text{soft}}$, we define the final target matrix as

\begin{equation}
T^{(m \to n)}
=
(1-\alpha) I
+
\alpha T^{(m \to n)}_{\text{soft}},
\end{equation}

where the parameter $\alpha \in [0,1]$ controls the mixing between soft-targets and the hard identity targets.

The row-wise soft cross-entropy loss is

\begin{equation}
\ell^{(m \to n)}_b
=
-
\sum_{k=1}^{B}
T^{(m \to n)}_{b,k}
\log
P^{(m \to n)}_{b,k}.
\end{equation}

The directional loss is

\begin{equation}
\mathcal{L}^{(m \to n)}
=
\frac{1}{B}
\sum_{b=1}^{B}
\ell^{(m \to n)}_b.
\end{equation}

\subsection{Weighting Faulty Positives}
The principle of our semantic weighting is demonstrated in Figure~\ref{fig:concept}.  
Positive pairs correspond to diagonal similarities

\begin{equation}
s^{(m,n)}_b
=
S^{(m,n)}_{b,b}.
\end{equation}

Let $\mu$ and $\sigma$ denote the mean and standard deviation
of the batch positive similarities. To stabilize these statistics over training, we follow~\cite{morgado2021audiovisual, wu2018unsupervised} and compute exponential moving averages.

We compute normalized scores

\begin{equation}
z_b
=
\frac{
s^{(m,n)}_b - (\mu + \delta \sigma)
}{
\sigma \sqrt{\kappa}
},
\end{equation}

where the parameters $\delta, \kappa > 0$ control the weight distribution.

Weights are obtained via the Gaussian cumulative distribution function

\begin{equation}
\Phi(z)
=
\frac12
\left(
1 +
\operatorname{erf}\left(
\frac{z}{\sqrt{2}}
\right)
\right),
\end{equation}

and

\begin{equation}
w_b
=
\Phi(z_b)(1 - w_{\min}) + w_{\min},
\end{equation}

where $w_{\min} \in [0,1]$ denotes the minimum positive weight.

The weighted directional loss is

\begin{equation}
\widetilde{\mathcal{L}}^{(m \to n)}
=
\frac{
\sum_{b=1}^{B}
w_b \ell^{(m \to n)}_b
}{
\sum_{b=1}^{B}
w_b
}.
\end{equation}

The bidirectional pair loss is

\begin{equation}
\widetilde{\mathcal{L}}^{(m,n)}
=
\eta \widetilde{\mathcal{L}}^{(m \to n)}
+
(1-\eta)
\widetilde{\mathcal{L}}^{(n \to m)},
\end{equation}

where the parameter $\eta \in [0,1]$ balances bidirectional losses.

\subsection{Multi-modal Objective}

Finally, the overall multi-modal loss averages over all unordered
modality pairs:

\begin{equation}
\mathcal{L}_{\text{GMA,RxID}}
=
\frac{2}{M(M-1)}
\sum_{1 \le m < n \le M}
\widetilde{\mathcal{L}}^{(m,n)}.
\end{equation}

This formulation jointly mitigates faulty negatives via soft targets
and faulty positives via similarity-based weighting, resulting
in robust global multi-modal alignment across all modalities.

\subsection{Modality-specific Encoders}

We follow CMVRA~\cite{cmvra} in using pretrained modality-specific encoders. 
For controlled evaluation of our multi-modal robust xID loss~\cite{morgado2021robust}, we adopt CLIP-ViP~\cite{CLIP_ViP} for RGB and skeleton as well as OMNIVORE~\cite{OMNIVORE} for IR and depth modalities, 
ensuring the performance gains stem from the novel loss rather than encoder variations.
    \section{Experiments}

In this study, we evaluate different loss functions for multi-modal representation learning on the downstream task of driver distraction detection. Therefore, we use LMA and GMA methods for self-supervised pretraining as well as linear evaluation on the Drive\&Act dataset~\cite{martin2019drive_and_act_2019_iccv} for evaluation.

\paragraph*{Training Details}For our experiments we use CLIP-VIP\cite{CLIP_ViP} and OMNIVORE~\cite{OMNIVORE} encoders. We modify the encoders to produce
512-dimensional embeddings, to enable comparison
between the different architectures within our self-supervised alignment framework.
We conduct our alignment using the AdamW optimizer
with a learning rate of 1e-4 and a batch size of 32. In order to meet the requirements of our encoder architectures, all video clips are represented by 12 uniformly sampled frames. For temporal consistency, we use the same sampling across all modalities. 

We train the models for 100 epochs. In order to stabilize the representation learning we start training using the InfoNCE loss for 30 epochs. Thereafter we use a linear scheduler to increase the $\alpha$ hyperparameter for 40 epochs from 0 to 0.5.

We set the temperature for our loss to 0.07 following other SotA methods~\cite{chen2024vision_language_skeleton, morgado2021robust}. For our loss function construction we use 0.5 for the weighting parameter $\eta$ and 0.25 for the minimum weight $w_{\min}$. We use data augmentation techniques including random center cropping with a range of 0.8-1.0 and horizontal flipping with a probability of 0.5.

\subsection{Dataset}

For our experiments, we use the multi-modal driver activity recognition dataset Drive\&Act~\cite{martin2019drive_and_act_2019_iccv} for both training and testing. Drive\&Act provides 34 fine-grained activity classes for driver secondary tasks making it suitable for our mutli-modal driver distraction detection. The Drive\&Act dataset consists of over 9.6 million frames recorded from six different cameras. The modalities include RGB, infrared, and depth data. In our experiments we leverage the four available modalities for our alignment and the six different IR views for cross-view evaluation. The challenging domain of in-cabin monitoring and the highly imbalanced data makes the dataset suitable for our proposed method. The test accuracies are reported as the mean balanced accuracy across the three available test splits of Drive\&Act.

\begin{table}[t]
\centering
\caption{Average balanced accuracy in [\%] for several baseline approaches and our GMA methods. Best results per modality are highlighted in bold.}
\label{tab:mean_bal_acc}
\begin{tabular}{l c c c c c}
\hline
\textbf{Loss} & \textbf{Alignment} & \textbf{RGB} & \textbf{IR} & \textbf{D} & \textbf{S} \\
\hline
InfoNCE~\cite{infoNCE} & LMA & \multirow{5}{*}{45.62} & 52.42 & 50.71 & 36.03   \\
WxID~\cite{morgado2021robust} & LMA            & & 55.51 & 50.89 & 36.03   \\
CMA-ST~\cite{chen2024vision_language_skeleton} & LMA  & & 57.16 & 51.71 & 35.70 \\
RxID~\cite{morgado2021robust} & LMA & & 51.95 & 49.57 & 39.86 \\
CMVRA~\cite{cmvra} & GMA & & 56.37 & 53.13 & 39.19 \\
\hline  
MM WxID & GMA & \multirow{3}{*}{45.62} & 56.57 & 53.36 & 37.77 \\
MM CMA-ST & GMA & & 55.38 & 53.31 & 38.43 \\
MM RxID    & GMA & & \textbf{57.30} & \textbf{54.25} & \textbf{40.72} \\
\hline
\textit{InfoNCE balanced} & \textit{GMA} & \textit{45.62} & \textit{58.92} & \textit{54.04} & \textit{38.28} \\
\hline
\end{tabular}
\label{tab:main_results}
\end{table}

\subsection{Evaluation Protocol}

In our alignment method, no labels are required. In order to quantify the capabilities of the learned representations, we evaluate the efficacy of our aligned encoders using the linear evaluation protocol (LEP) as proposed by Zhang et al.~\cite{zhang2016ColorfulIC}. The LEP is a widely used method for evaluating representations that have been learned without labels.
In accordance with the LEP, the encoders are initially aligned in a self-supervised manner without the utilization of any labeled data. Subsequently, the weights of the aligned encoders are frozen. We add a randomly initialized linear layer on top of the frozen encoder and train the linear layer supervised with labeled data. With this procedure, we can quantify the quality of the learned representations and report classification accuracies on the test sets.

    \begin{table}[t]
\centering
\small
\caption{Average balanced accuracy in~[\%] for cross-dataset evaluation on all IR views of Drive\&Act~\cite{martin2019drive_and_act_2019_iccv}. 
Best results on novel views are highlighted in bold.}
\setlength{\tabcolsep}{3.5pt}
\scalebox{0.9}{
\begin{tabular}{lcccccc}
\toprule
& \multicolumn{6}{c}{\textbf{Trained on Kinect IR}} \\
\midrule

\shortstack{Method} 
& \shortstack{I3D \\ \cite{carreira2017quo}}
& \shortstack{CMA-ST \\ \cite{chen2024vision_language_skeleton}}
& \shortstack{INCE \\ \cite{infoNCE}} 
& \shortstack{RxID \\ \cite{morgado2021robust}} 
& \shortstack{CMVRA \\ \cite{cmvra}} 
& \shortstack{MM-RxID \\ Ours}\\

\midrule
\textbf{Evaluated on} &&&&&&\\
\midrule

Kinect IR 
& 72.90 & 57.16
& 52.42 & 51.95
& 55.38 & 57.30 \\

\midrule

Inner Mirror 
& 6.60 & 37.90 
& 40.13 & 35.48 
& 39.36 & \textbf{41.38} \\

Wheel 
& 4.27 & 22.79
& 20.07 & 19.36
& \textbf{23.01} & 22.84 \\

Driver 
& 9.02 & 41.19
& 36.14 & 35.11
& 38.87 & \textbf{40.09} \\

Co Driver 
& 19.79 & \textbf{48.38}
& 43.30 & 42.65
& 46.48 & 46.40 \\

Ceiling 
& 7.34 & 34.54
& 35.09 & 32.39
& 34.35 & \textbf{35.57} \\

\midrule

Avg (unseen)
& 9.40 & 32.84
& 34.94 & 33.00
& 36.41 & \textbf{37.26} \\

\bottomrule
\end{tabular}}
\label{tab:crossvw_full}
\end{table}

\section{Results}
\subsection{Multi-modal Alignment}

Table~\ref{tab:main_results} shows the mean balanced accuracy for each modality averaged over dataset splits, comparing our proposed robust global alignment to traditional LMA and the GMA baseline CMVRA~\cite{cmvra}.
For LMA, standard InfoNCE~\cite{infoNCE} provides a reasonable baseline, with moderate performance across IR, Depth, and Skeleton modalities. The robust variant weighted xID (WxID)~\cite{morgado2021robust} outperforms the InfoNCE or is on par for all modalities. RxID~\cite{morgado2021robust} improves the performance on skeleton but falls short in IR and depth modalities compared to InfoNCE. The LMA using soft targets as proposed by~\cite{chen2024vision_language_skeleton} (CMA-ST) outperforms the InfoNCE baseline on IR and depth but falls short on skeleton modality.

In the GMA setting, CMVRA demonstrates that extending alignment to all modality pairs can improve performance over pairwise methods in all available modalities. Our proposed multi-modal extensions MM-WxID and MM-CMA-ST show mixed improvement compared to CMVRA. However, MM-RxID further increases accuracy compared to CMVRA across all evaluated modalities. Notably, MM-Robust xID achieves the highest balanced accuracy on IR, Depth, and Skeleton, with gains of approximately $+0.9\%$ up to $+2.3\%$ over CMVRA. These results confirm that jointly addressing faulty negatives (via soft targets) and unreliable positives (via weighting) in a global multi-modal framework leads to robust, consistently superior representations.

We also report results for alignment on a balanced  version of Drive\&Act. The results are reported as \textit{InfoNCE balanced} and provide an upper bound reference, illustrating the potential for further improvement when combining global alignment with robust losses. Overall, the results demonstrate that our weighted soft-target approach scales effectively to multiple modalities while consistently outperforming both LMA and existing GMA baselines.

These findings underscore the relevance of a holistic GMA. The utilization of RxID alone in the LMA setting does not result in a discernible enhancement in performance metrics. It has been demonstrated that the most effective approach to achieving robust multi-modal representation alignment is to combine RxID with GMA. 

\subsection{Cross-view Generalization Ablation}

To evaluate the generalization of our method to novel views, we conducted cross-view experiments on the Drive\&Act dataset~\cite{martin2019drive_and_act_2019_iccv}, training on Kinect IR and evaluating on retained camera views. For CMVRA and MM-RxID, we align the IR encoder to all other modalities and evaluate the frozen IR encoder which has only seen IR train images. Table~\ref{tab:crossvw_full} reports the average balanced accuracy across all views and the mean accuracy over all views.
Several observations emerge from these results. First, standard LMA using either InfoNCE, CMA-ST or RxID achieves reasonable performance on unseen views, highlighting the baseline ability of local instance discrimination to generalize. Compared to the supervised trained I3D~\cite{carreira2017quo} baseline, reported by Martin et al.~\cite{martin2023viewpoint}, self-supervised learning shows superior performance in view generalization. However, the performance on LMA is limited, particularly on extreme novel views such as ceiling and steering wheel.

Second, our proposed GMA improves generalization across almost all novel views. Specifically, GMA with our MM-RxID yields the best results on inner mirror, a-column driver, and ceiling views, with gains of up to $+5.9\%$ over the corresponding LMA baseline.
We find that for InfoNCE the GMA doesn't consistently yield better generalization than LMA, specifically on the inner mirror and ceiling views.
The a-column co-driver camera view angle is quite close to the trained Kinect IR view. CMA-ST performs best on the co-driver view and second best on the trained Kinect IR view. 

Overall, these ablations underscore that our weighted soft-target approach for GMA not only enhances local instance discrimination but also substantially improves generalization to novel camera perspectives, a key requirement for robust multi-view understanding in real-world scenarios.

\subsection{Discussion}

The results in Tables~\ref{tab:main_results} and~\ref{tab:crossvw_full} highlight several key insights regarding the effectiveness and limitations of our proposed multi-modal alignment framework.
Comparing LMA and GMA, we observe that extending alignment beyond pairwise interactions consistently improves performance across all modalities. For instance, the GMA variants CMVRA, MM-WxID and MM-RxID consistently outperform their LMA counterparts, particularly on challenging modalities such as Depth and Skeleton. This finding indicates that by aggregating alignment across all modality pairs, the model is able to acquire more robust and coherent representations, thereby reducing the risk of overfitting to single-modal information.

Among the different losses, MM-RxID achieves the best overall performance, highlighting the importance of jointly addressing noisy positives and faulty negatives. MM-WxID provides some improvement over CMVRA. But MM-RxID delivers additional gains, indicating that handling the hard-negative assumption is critical for global alignment especially on the challenging skeleton modality. The MM-CMA-ST yield modest improvements suggesting that the original loss, designed for vision and language, is not suited for aligning multiple vision modalities. 

The results demonstrate the efficacy of weighting in a more extensive context. The findings indicate that a multitude of distinct data modalities, each comprising complementary information, are optimally suited for this configuration. This development enables the exploration of non-image-related sensor modalities, including audio and physiological data, such as EEG signals, which are not typically associated with visual perception.

The cross-view ablations in Table~\ref{tab:crossvw_full} show that our method generalizes better to unseen camera views. GMA with robust losses consistently outperforms LMA, reducing variance across views and achieving the best results on most novel perspectives. These findings indicate that both global alignment and robustness mechanisms contribute to generalization, with the combination of soft targets and positive reweighting particularly effective for handling semantic overlap and viewpoint changes.

Despite the improvements, certain modalities remain challenging, where gains from robust losses are limited. The performance of our MM-RxID compared to the \textit{balanced InfoNCE} demonstrates the strength of our approach in lifting difficult modalities like skeleton. Conversely, it also shows the trade-off in the performance drop in the IR modality. Future work should address this limitation and introduce modality specific weighting e.g. with an entropy-based weighting. Additionally, while the weighted soft-target approach improves generalization, it introduces additional hyperparameters (\(\delta, \kappa, w_{\min}\)) and computational overhead, which may require careful tuning for larger modality counts. 
Overall, our experiments support the hypothesis that robust GMA is necessary for scalable and generalizable representation learning in driver distraction detection. The ablation studies underline that both positive reweighting and soft-target modeling contribute meaningfully, and that their combination in the MM-RxID framework consistently achieves the strongest performance across both seen and novel views.
    \section{Conclusion}

We present a robust framework of GMA for driver distraction detection. In our work, we extend traditional pairwise alignment and propose GMA which aligns all modalities jointly. We extend pairwise robust xID loss which addresses the challenges of faulty negatives and unreliable positives to a global MM-RxID loss.  
Extensive experiments on the Drive\&Act dataset demonstrate that our proposed MM-RxID consistently outperforms both traditional pairwise alignment including the RxID and the existing GMA method CMVRA across all modalities. Cross-view ablations further show that our approach improves generalization to unseen perspectives, reducing performance variance across camera views and enhancing robustness in real-world multi-modal scenarios.

The findings of this study underscore the conclusion that the utilization of RxID alone, employing LMA, does not result in a discernible enhancement in performance metrics. It has been demonstrated that the most effective approach to achieving robust multi-modal representation alignment is to combine RxID with GMA.
Our proposed method is extendable to more modalities including 1-D data such as audio and EEG signals.
In summary, our work provides a scalable and generalizable solution for multi-modal representation learning, paving the way for more reliable cross-modal understanding in driver monitoring applications. Future work may explore extending these ideas to larger modality counts and more diverse real-world datasets, as well as investigating modality adaptive weighting strategies to further improve robustness.

\section*{Acknoledgement}
The work has been funded under the funding code 19A24002K by the Federal Ministry for Economic Affairs and Energy of Germany (BMWE) on the basis of a decision by the German Bundestag and by the European Union. The work was performed in the project SALSA (https://projekt-salsa.de/).
	
	\bibliographystyle{IEEEtran}
	\bibliography{root} 
	
\end{document}